\documentclass[runningheads]{llncs}
\usepackage{graphicx}
\usepackage{booktabs}
\usepackage{tabularx}
\usepackage{textgreek}
\usepackage{multirow}
\usepackage{float}

\begin{document}
\title{Integrating Large Language Models and Knowledge Graphs for Extraction and Validation of Textual Test Data}
\author{Antonio De Santis\inst{1}\and
Marco Balduini\inst{2,3}\and
Federico De Santis\inst{3}\and
Andrea Proia\inst{4} \and 
Arsenio Leo\inst{4}\and 
Marco Brambilla\inst{1} \and
Emanuele Della Valle\inst{1}
}

\authorrunning{A. De Santis et al.}
%
\titlerunning{Integrating LLMs and KGs for Extraction and Validation of Test Data}
\institute{Politecnico di Milano, DEIB, I-20133 Milano, Italy \\
\email{\{firstname.lastname\}@polimi.it} \and 
Quantia Consulting, Milano, Italy \\
\email{marco.balduini@quantiaconsulting.com}\and 
motus ml, Milano, Italy \\
\email{\{firstname.lastname\}@motusml.com}\and 
Thales Alenia Space, Roma, Italy \\
\email{\{firstname.lastname\}@thalesaleniaspace.com}}

\maketitle              
\begin{abstract}
Aerospace manufacturing companies, such as Thales Alenia Space, design, develop, integrate, verify, and validate products characterized by high complexity and low volume. They carefully document all phases for each product but analyses across products are challenging due to the heterogeneity and unstructured nature of the data in documents. In this paper, we propose a hybrid methodology that leverages Knowledge Graphs (KGs) in conjunction with Large Language Models (LLMs) to extract and validate data contained in these documents. We consider a case study focused on test data related to electronic boards for satellites. To do so, we extend the Semantic Sensor Network ontology. 
We store the metadata of the reports in a KG, while the actual test results are stored in parquet accessible via a Virtual Knowledge Graph. The validation process is managed using an LLM-based approach. We also conduct a benchmarking study to evaluate the performance of state-of-the-art LLMs in executing this task. Finally, we analyze the costs and benefits of automating preexisting processes of manual data extraction and validation for subsequent cross-report analyses.

\end{abstract}
\section{Introduction}
\textbf{Context. }
Companies in the aerospace industry produce complex products in low volumes. As a result, most of the data that can boost analytics is hidden within documents, making its extraction challenging.
The experience presented in this article focuses on Test Data related to electronic boards used in Thales Alenia Space's satellite systems. The production of these electronic boards is a critical aspect of space technology~\cite{norman}. These boards are manufactured in limited quantities, with a satellite containing between 10 to 20 such boards. Moreover, these components must be extremely reliable and are subject to rigorous testing protocols due to the hostile conditions of space missions~\cite{ghedini}. Given the near impossibility of conducting repairs once satellites are in space, production errors could potentially lead to the failure of an entire mission, which would result in significant financial losses and wasted resources. 
In this scenario, data analytics can play a crucial role, providing timely insights and enabling immediate actions based on the data's flow and characteristics. For example, the analysis of historical production data could reveal trends that can predict the likelihood of future components failing the quality tests. Such insights can guide production decisions, minimizing waste and resulting in significant cost savings.

\vspace{0.25cm}\noindent\textbf{Problem Statement. }
The effectiveness of these data-driven approaches relies on the quality and organization of the data~\cite{data_quality}. Each electronic board is meticulously crafted and tested before receiving approval. However, the testing procedures and the generation of Test Reports are manually executed by human operators across multiple isolated documents (primarily in .docx and .pdf format). 
This leads to data that is highly fragmented, heterogeneous, unstructured, and prone to errors and inconsistencies. Such a scenario poses a significant challenge, as it can jeopardize data analysis efforts. Considering this, the focus of our case study is automating the extraction, validation, and integration of Test Data. Given the high level of data heterogeneity, the process of validation is particularly challenging because a standard approach based on regular expressions would be impractical.

\vspace{0.25cm}\noindent\textbf{Proposed Solution. }
To address the aforementioned challenges, we propose a hybrid approach that utilizes Large Language Models (LLMs) in combination with Semantic Web technologies. 
To provide semantic knowledge to the system and manage structural heterogeneity, we create an ontology to capture the semantics of the data. This ontology extends the \emph{Semantic Sensor Network} (SSN)~\cite{SSN} ontology, a well-established ontology for representing sensor data.
We then proceed with extracting the data from Test Report documents and storing it in tabular format. The extracted data must undergo an automatic validation process (i.e., checking for inconsistencies in test results). For this task, we exploit the implicit knowledge of LLMs. These models have demonstrated their capability to process data despite structural and syntactic heterogeneity. 
Moreover, in contrast with approaches based on regular expressions, LLMs have the advantage of being able to scale effectively with an increasing variety and complexity of the data.
The validated data is integrated into a data storage system, ensuring a structured and organized data repository. To facilitate direct data access, we then create mappings between the data storage and the ontology, allowing our system to understand the relationships and connections among data points. This knowledge is stored in a Virtual Knowledge Graph (VKG)~\cite{vkg}, also known in the literature as Ontology-Based Data Access (OBDA)~\cite{obda}, and is accessed using SPARQL queries, which are automatically translated into SQL language. 

\vspace{0.25cm}\noindent\textbf{Structure of the Work. }
The paper is structured as follows. Section~\ref{related} presents a review of related work. Section~\ref{case} describes the case study in detail and Section~\ref{motivation} explains the rationale behind using KGs and LLMs. Section~\ref{methodology} presents our methodology, whose implementation and evaluation are detailed in Section~\ref{impl}. Section~\ref{lessons} discusses the uptake of our work and the lessons learned, while Section~\ref{conclusions} concludes the paper, providing directions for future work.

\section{Related Work}
\label{related}
\textbf{Industrial deployment of VKGs. }
Semantic Web technologies have been successfully applied in several industrial contexts~\cite{railway,renault,ecommerce} as they simplify data access by providing an abstraction layer (i.e., an ontology) that integrates data from semantically and physically different sources. Siemens uses an OBDA for managing the temperature data of trains and turbines and developed a semantic rule-based diagnostic system~\cite{iot,siemens2,siemens3}. Statoil has implemented an OBDA using the \emph{Ontop}~\cite{ontop} framework for integrating multiple data sources~\cite{statoil}. This system has enhanced the efficiency of data collection for geologists in the field of oil and gas exploration and production. Similarly, Ontop was used to realize a semantic information model for managing machine data~\cite{manufacturing}. Moreover, Ford Motor Company also stores knowledge about manufacturing processes in an ontology~\cite{ford}. This allows their internally developed AI system to handle the planning of vehicle assembly processes. They have also explored the use of federated ontologies to identify potential risks in the supply chain~\cite{ford2,ford3}. Bosch also has utilized ontology-based approaches for data access. They applied knowledge graph embedding~\cite{bosch} and ontology reshaping~\cite{bosch2} for automatic knowledge graph (KG) construction in a case related to welding quality monitoring.

\vspace{0.25cm}\noindent\textbf{LLMs for Data Management. }
In recent years, the field of language models has experienced substantial progress due to the introduction of LLMs such as GPT-3.5~\cite{gpt3} and GPT-4~\cite{gpt4}, developed by OpenAI, Meta's Llama 1~\cite{llama1} and Llama 2~\cite{llama2}, Claude 3~\cite{claude3} from Anthropic, Google's Gemini~\cite{gemini}, and Mixtral~\cite{mixtral}, from Mistral AI. These models have been utilized in a variety of data management tasks~\cite{dbgpt,llmenhanced} due to their ability to extract knowledge from unstructured data sources~\cite{knowledgeeng} and to understand the data without the need for explicit modeling~\cite{disrupt}. From a data validation perspective, LLMs have demonstrated close to human-level capabilities in detecting inconsistencies in text summaries~\cite{inconsistencies}. 
In the context of the Semantic Web, LLMs can also be used to automate KG completion and construction~\cite{llmkg}. For instance, GPT-4 was used for automatic ontology and KG construction for large amounts of unstructured sustainability-related data~\cite{trajanoska2023enhancing}.
Moreover, LLMs have been effectively utilized to assist with data preparation tasks required before performing business analytics~\cite{businessLlms}. More specifically, GPT-4 was used to translate product names, assign product categories, classify customer sentiment, and extract repair requests and their causes from customer service logs.
Regarding real industry scenarios, there is currently limited evidence, to our knowledge, of LLMs being utilized in conjunction with semantic technologies for data validation in large manufacturing companies.

\section{Case Study: Testing of Electronic Boards}
\label{case}
In this section, we discuss our case study in greater detail and describe the structure and characteristics of Thales Alenia Space's Test Reports.

\begin{figure*}[t]
    \centering
    \includegraphics[width=0.99\textwidth]{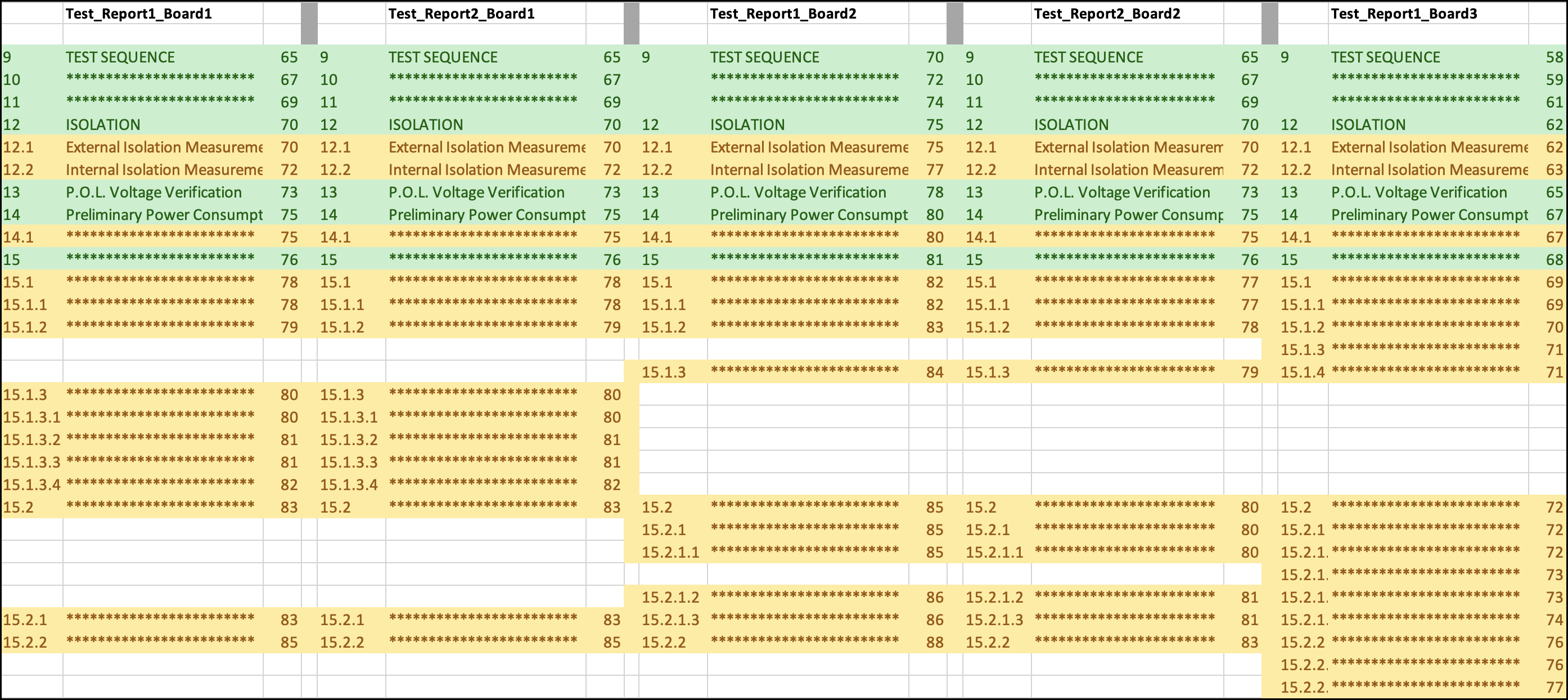}
    \caption{A portion of a color-coded spreadsheet that visually represents the heterogeneity within Test Reports, which typically contain around 23 sections. Green denotes uniform sections, while yellow represents variable ones. White cells indicate the absence of a section. Titles are intentionally obscured to protect confidential information.}
    \label{fig:heatmap}
\end{figure*}

\begin{figure*}[t]
    \centering
    \includegraphics[width=1\textwidth]{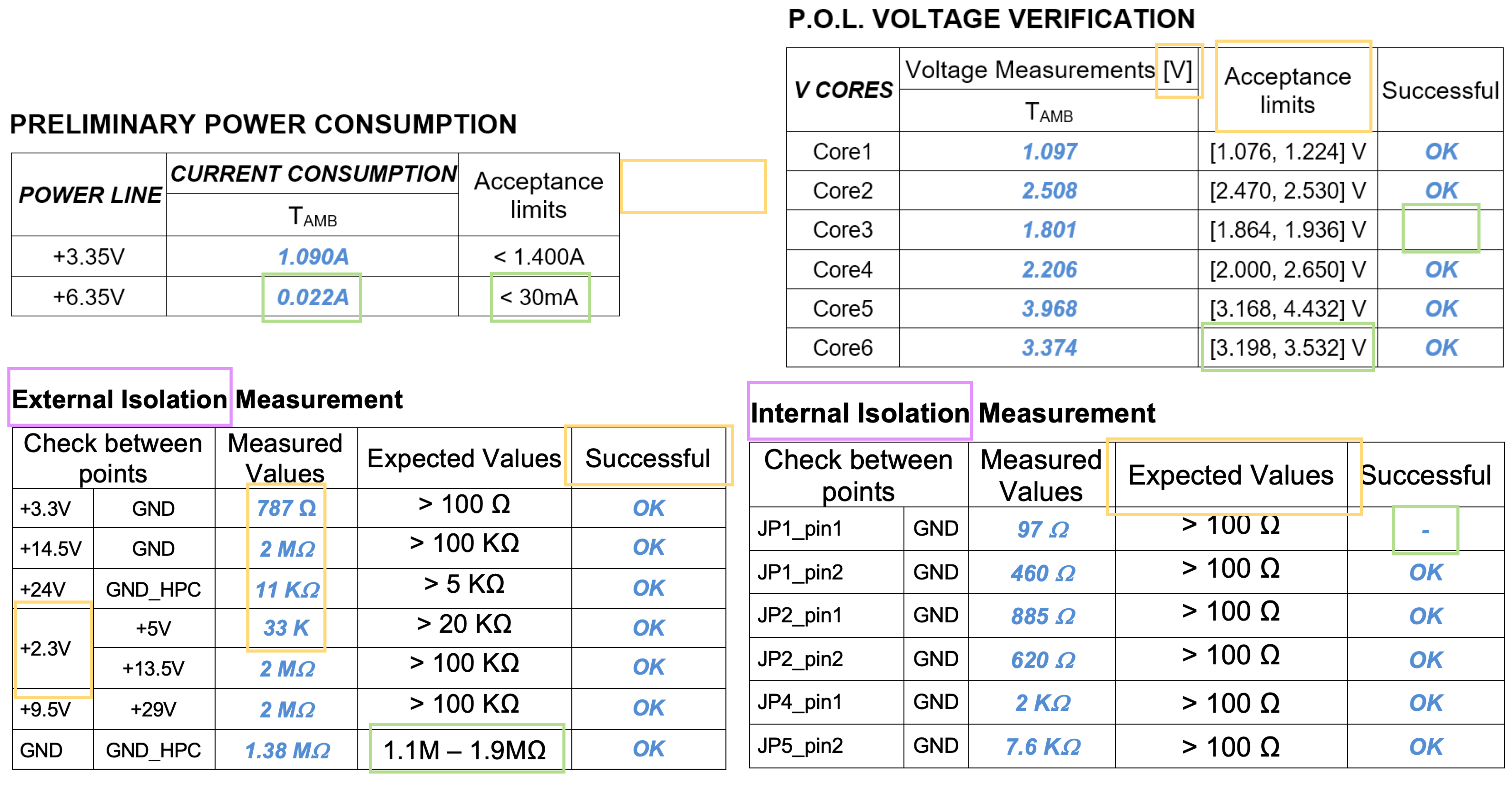}
    \caption{Examples of test results tables that illustrate the challenges of syntactic (shown in green), structural (shown in yellow), and semantic (shown in purple) heterogeneity.}
    \label{fig:tables}
\end{figure*}

\vspace{0.25cm}\noindent\textbf{Electronic Boards Test Data. }
Our case study involves Test Data for electronic boards, primarily Printed Circuit Boards (PCBs) used in satellite systems. Testing these products is a critical process in the space industry, ensuring that all technological processes meet specific mission requirements and comply with standards established by the European Space Agency (ESA) and the European Cooperation for Space Standardization (ECSS). The tests involve measuring parameters such as voltage, resistance, or power, and comparing the results to a predefined expected range, which represents the acceptable limits within which the parameter should fall for the PCB to operate correctly. Test engineers conduct these tests, which are documented in Test Reports. These documents, which are primarily in .docx and .pdf format, are manually filled by the engineers and exhibit a high degree of heterogeneity. In Figure~\ref{fig:heatmap}, we show a color-coded spreadsheet to illustrate the heterogeneity within these documents.
The actual test results in the reports are organized within manually filled tables. The “acceptance limits” column is pre-filled and the engineers have to fill in the measured value and a “successful” column based on the test outcome. In this study, we consider Point-of-Load (POL) Voltage Verification, Preliminary Power Consumption, and Isolation (both external and internal) as representative types of tests. Figure~\ref{fig:tables} provides an example of tables for these types of tests, emphasizing the unstructured and heterogeneous nature of the data which manifests in several ways:

\vspace{0.1cm}\noindent\emph{Syntactic Heterogeneity. } This is seen in the different formatting of the data. The range of acceptance limits is represented in various ways. For instance, “[3.198, 3.532] V" and “1.1M - 1.9M\textOmega " both indicate a range of acceptance. 
In some cases, the measured value and the acceptance limits are indicated with different units of measure.
Additionally, in the “successful" column, the absence of a value or the presence of a “-" both indicate a lack of success.

\vspace{0.05cm}\noindent\emph{Structural Heterogeneity. } This is evident in the inconsistent organization and naming of the tables. For instance, some tables have a single “successful" cell in a different part of the document and therefore lack a dedicated “successful" column. Furthermore, a column labeled “Acceptance limits" in one table might be labeled as “Expected Values" in another. 
The unit of measure can be included in the table title as well as written with the values or even absent. Another form of structural heterogeneity can be observed in the use of row span, which is used to indicate that the same value applies to multiple rows. 

\vspace{0.05cm}\noindent\emph{Semantic Heterogeneity}. There is an implicit hierarchical structure within the reports as there are various representations for the concept of a “Test", such as “Internal Isolation", “External Isolation" or “POL Voltage". These test types share many properties, but they are categorized separately due to their specific aspects. Similarly, Internal and External Isolation fall under the category of Isolation tests, each possessing properties specific to Isolation testing. Despite this, they are represented differently, introducing a semantic heterogeneity. This leads to the requirement of modeling what is a “Test” or an “Isolation test”.

\vspace{0.05cm}\noindent The preexisting manual approach (see Figure~\ref{fig:pipeline_as_is}) for data extraction and validation is costly, time-consuming, and allows for limited cross-report analyses, but automating these processes isn't straightforward. Although a human operator can intuitively understand that, for example, “Acceptance limits" has the same meaning as “Expected values", this poses a challenge for an automated system.

\renewcommand{\thefigure}{3}
\begin{figure*}[t]
\centering
\includegraphics[width=0.98\textwidth]{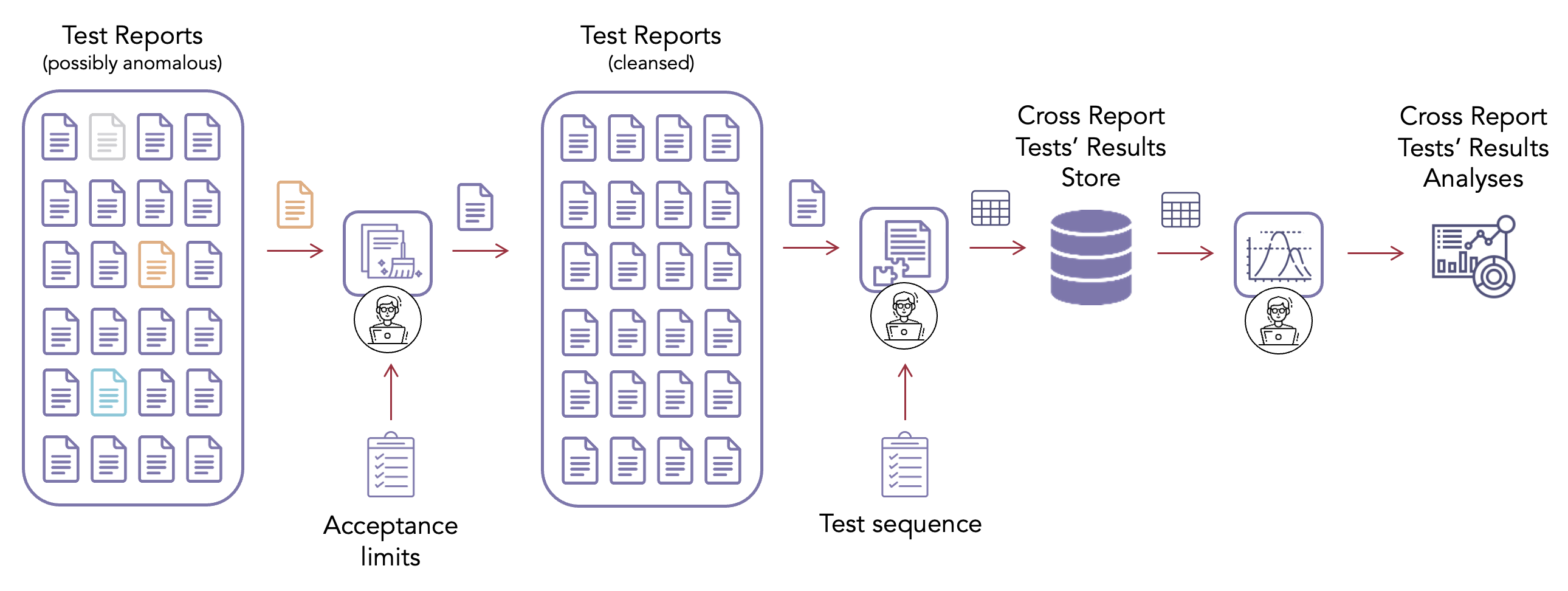}
    \caption{The preexisting manual data processing workflow in which potentially anomalous reports are subjected to manual extraction, cleaning, and validation.}
    \label{fig:pipeline_as_is}
\end{figure*}

\vspace{0.25cm}\noindent\textbf{Data Obfuscation. }
Data is not disclosed in its original form to protect Thales Alenia Space's privacy. We added noise to the values, ensuring the structure and syntax remained intact without disclosing any confidential information.

\section{Motivation}
\label{motivation}
In this section, we aim to clarify our motivation by addressing two key questions: (1) Why do we need KGs? and (2) Why integrate them with LLMs?

\vspace{0.25cm}\noindent\textbf{Motivation for Knowledge Graphs. }
The motivation for choosing KGs and OBDA systems lies in their ability to handle heterogeneous and physically distributed information, a common challenge in knowledge-intensive industries such as aerospace.
KGs effectively accommodate the high diversity and low volume of data in the space industry, which produces hundreds of PCB families (with similar but not identical designs) but only a few dozen PCBs. The industry also deals with a diverse array of tests due to the intricate nature of PCBs, which include passive and active electrical components, as well as digital electronics like RAM, CPUs, and FPGAs.
Leveraging and extending resources such as the SSN ontology can facilitate the modeling process in this case. Furthermore, a graph-based representation allows for a more explicit data repository, reducing the reliance on tacit knowledge held by domain experts. This is crucial in aerospace where semantic coherence is key for managing complex systems such as satellites.

\vspace{0.25cm}\noindent\textbf{Integrating LLMs with KGs. }
Consider the detailed RDF representation in Listing~\ref{list:complex} that includes the QUDT (Quantities, Units, Dimensions, and Types)~\cite{QUDT} ontology for the units of measurement. Annotating data in this way would require a large amount of manual work at the level of the template of the Test Report. This can be challenging and time-consuming when dealing with complex and diverse data. Moreover, the complexity grows with the number of different templates of Test Reports the company introduces (i.e., one per PCB family). See once again Figure~\ref{fig:heatmap} to feel the degree of heterogeneity at the level of the sections of the reports.
However, LLM’s ability in natural language understanding can determine whether a measured value falls within an expected range, even if the syntax changes or the units of measurement differ. Therefore, it can assist in error detection and simplify the modeling process. This leads to a lightweight annotation of the data (see Listing~\ref{list:simple}) using the Unified Code for Units of Measure (UCUM)~\cite{ucum}, allowing data engineers to focus on the conceptual model and semantic meaning of the data, without having to account for every minor syntactic heterogeneity.

\renewcommand{\figurename}{Listing}
\begin{figure}
\renewcommand{\thefigure}{1.1}
\begin{verbatim}
<http://tasi.com/pol#TASI-1234-Core1> a sosa:Observation ;
    rdfs:label "TASI-1234-Core1" ;
    sosa:observedProperty tasi:POLVoltage ;
    sosa:hasResult [
      a qudt-1-1:QuantityValue ;
      qudt-1-1:numericValue "1.097"^^xsd:double ;
      qudt-1-1:unit qudt-unit-1-1:Volt 
   ] ;
   tasi:hasAcceptanceLimits [
      a tasi:Range ;
      tasi:lowerLimit [
         a qudt-1-1:QuantityValue ;
         qudt-1-1:numericValue "1.076"^^xsd:double ;
         qudt-1-1:unit qudt-unit-1-1:Volt 
      ] ;
      tasi:upperLimit [
         a qudt-1-1:QuantityValue ;
         qudt-1-1:numericValue "1.224"^^xsd:double ;
         qudt-1-1:unit qudt-unit-1-1:Volt 
      ] ;
   tasi:hasTestResult "OK" ;
   tasi:reportedIn "TASI-1234" ;
   tasi:testReportDate "2023-06-15"^^xsd:dateTime .
\end{verbatim}
\caption{Detailed representation that a machine can understand without LLMs.}
\label{list:complex}
\end{figure}

\vspace{-0.25cm}


\renewcommand{\thefigure}{1.2}
\begin{figure}[H]
\begin{verbatim}
<http://tasi.com/pol#TASI-1234-Core1> a sosa:Observation ;
    rdfs:label "TASI-1234-Core1" ;
    sosa:observedProperty tasi:POLVoltage ; 
    sosa:hasSimpleResult "1.097 V"^^cdt:ucum ;
    tasi:hasAcceptanceLimits "[1.076, 1.224] V" ;
    tasi:hasTestResult "OK" ;
    tasi:reportedIn "TASI-1234" ;
    tasi:testReportDate "2023-06-15"^^xsd:dateTime .
\end{verbatim}
\caption{Lightweight representation that can be understood by LLMs.}
\label{list:simple}
\end{figure}
\renewcommand{\figurename}{Fig.}

\renewcommand{\thefigure}{4}
\begin{figure*}[t]
    \centering
    \includegraphics[width=1\textwidth]{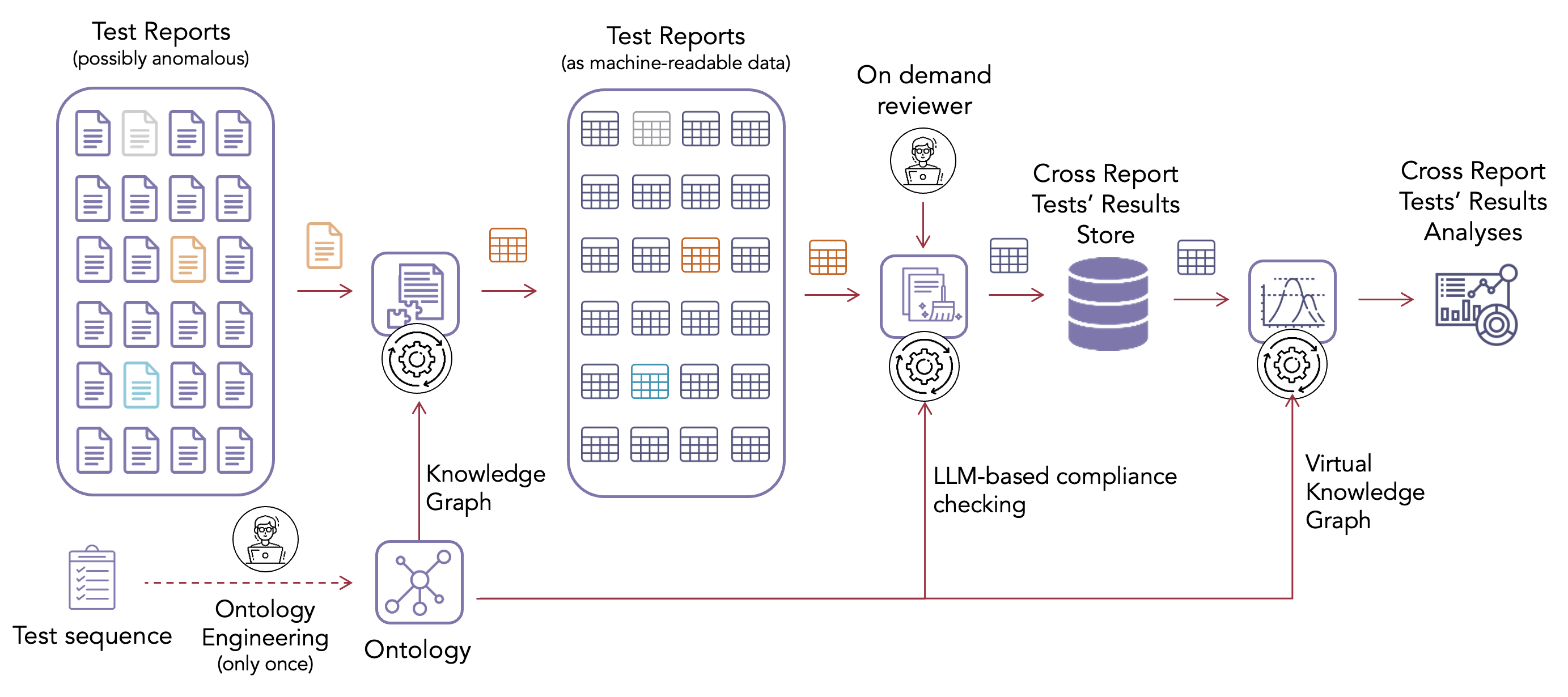}
    \caption{A flowchart representation of the proposed methodology. The process begins with the input of a set of potentially anomalous Test Reports, from which the data is extracted and transformed into a machine-readable format. The documents' metadata is integrated into a KG, while the test results undergo LLM-based compliance checking and anomalies are handled by an on-demand reviewer. The validated data is accessed through a VKG, enabling access to heterogeneous data and facilitating cross-report analyses. The whole process is guided by a one-time ontology engineering process.}
    \label{fig:pipeline}
\end{figure*}

\section{Methodology}
\label{methodology}
In this section, we describe the methodology of our approach for extraction, validation, and integration of Test Data from unstructured Test Reports. As depicted in Figure~\ref{fig:pipeline}, the process is divided into several phases. The validation process is managed using an LLM-based approach. On the other hand, data integration is accomplished through KGs, enabling access to heterogeneous data. More specifically, the process is structured on three levels:
\begin{itemize}
\item \textit{Data Extraction:} Test Reports' metadata and the test types they contain are extracted and stored in a KG using an ontology.
\item \textit{LLM-Based Compliance Checking:} LLMs are used to validate that test results are consistent with their respective acceptance limits.
\item \textit{Ontology-Based Data Access:} A VKG is used to mediate the actual access to the test results.
\end{itemize}

\renewcommand{\thefigure}{5}
\begin{figure*}[t]
    \centering
    \includegraphics[width=1\textwidth]{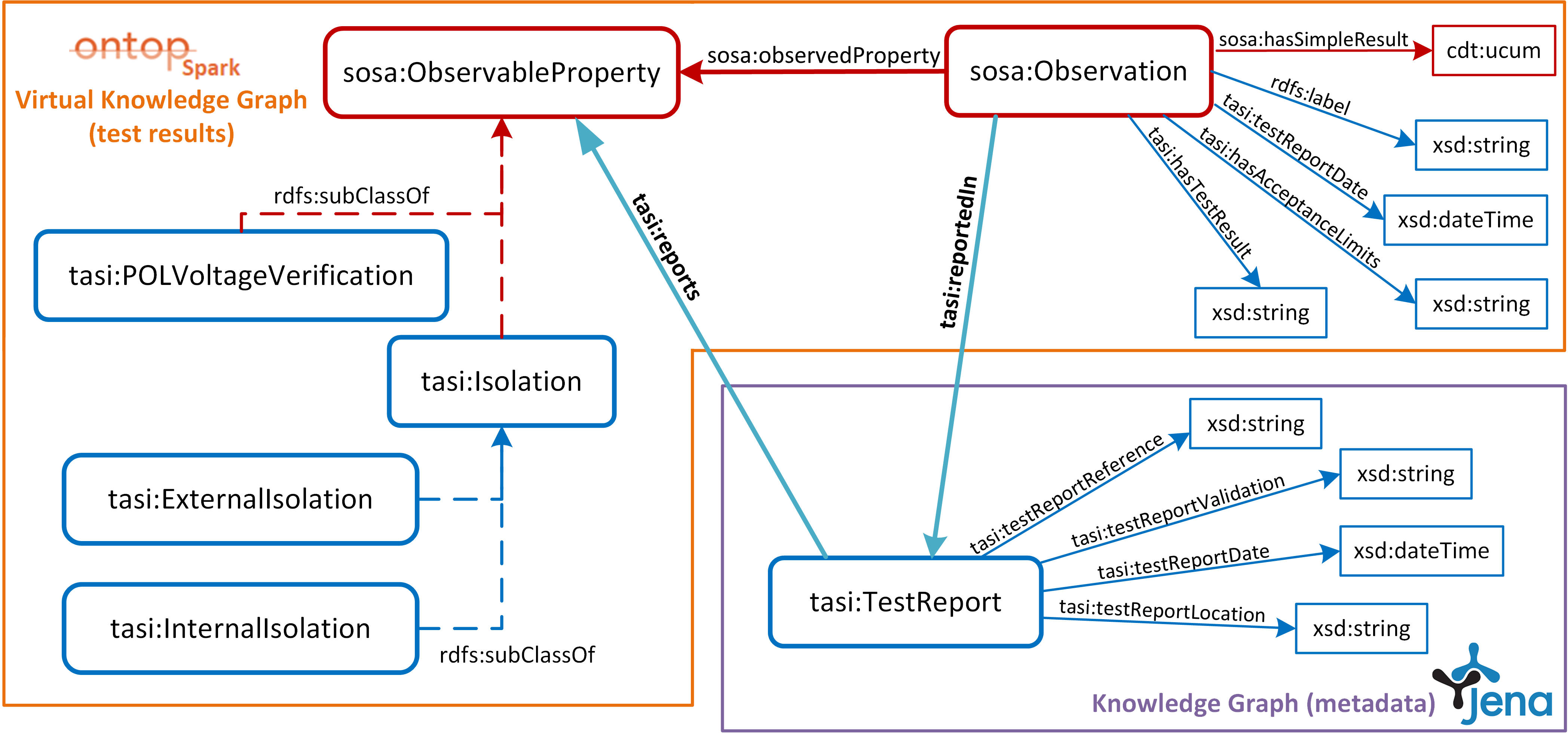}
    \caption{The figure provides a visual representation of our ontology. This ontology is an extension of the well-established SSN ontology, which is denoted by the \texttt{sosa} prefix and the red color. Our additions include new classes and properties, which are identified by the \texttt{tasi} prefix and the blue color.
    The ontology's components used for modeling Test report metadata are enclosed within a purple rectangle, and this metadata is stored within a KG. The modeling of test results, represented by an orange rectangle, is stored in a structured data repository and made accessible via a VKG.
    }
    \label{fig:ontology}
\end{figure*}

\vspace{0.25cm}\noindent\textbf{Data Extraction. }
The first step in our process is to extract the textual data within the Test Reports and transform it into a machine-readable format. This transformation is facilitated by a one-time ontology engineering process that defines the concepts, categories, and relationships embedded within the data. A KG is used for this purpose. The ontology used in this KG is an extension of the SSN ontology (see Figure~\ref{fig:ontology}) and maps the information related to Test Reports and observable properties found within these reports, which refer to the property being tested (i.e., the test type) and the related test table structure description in terms of its columns. All tests are of type \texttt{sosa:ObservableProperty} with their respective hierarchy. For instance, \texttt{<POLVoltage>} and \texttt{<Isolation>} are defined as \texttt{sosa:ObservableProperty}. \texttt{<InternalIsolation>} and \texttt{<ExternalIsolation>} are defined as sub-classes of \texttt{<Isolation>}. The RDF fragment provided in Listing~\ref{list:testreport} is an example of how a Test Report is modeled. An additional property \texttt{tasi:reports} has been added due to the absence of a Test Report concept in the SSN ontology.
\renewcommand{\figurename}{Listing}
\begin{figure}[h]
\renewcommand{\thefigure}{2}
\begin{verbatim}
@prefix tasi:   <http://www.semanticweb.org/ontologies/tasi#> .

tasi:POLVoltage a sosa:ObservableProperty ;
  rdfs:label "P.O.L. Voltage"@en .
  tasi:obsPropertyAccLimLocation  "VALIDATED/ist/TASI-1234-ist_pol.csv" ;
  tasi:obsPropertyResultsLocation "/VALIDATED/pol/pol.parquet" .
tasi:Isolation a sosa:ObservableProperty ;
  rdfs:label "Isolation"@en .
tasi:InternalIsolation rdfs:subClassOf tasi:Isolation ;
  rdfs:label "Internal Isolation"@en .
  tasi:obsPropertyAccLimLocation  "VALIDATED/ist/TASI-1234-ist_al.csv" ;
  tasi:obsPropertyResultsLocation "VALIDATED/ist/ist.parquet" .
tasi:ExternalIsolation rdfs:subClassOf tasi:Isolation ;
  rdfs:label "External Isolation"@en .
   
<http://tasi.com#TASI-1234> a tasi:TestReport ;
  tasi:reports tasi:POLVoltage, tasi:InternalIsolation ;
  tasi:testReportDate "2023-06-15"^^xsd:dateTime ;
  tasi:testReportName "test_report_xy" ;
  tasi:testReportReference "TASI-1234" ;
  tasi:testReportValidation "OK" ;
  tasi:testReportLocation "path/to/test_report_file.docx" .
\end{verbatim}
\caption{An example of a Test Report modeled using an extension of the SSN ontology that contains valid data for the \texttt{POLVoltage} and \texttt{InternalIsolation} tests.}
\label{list:testreport}
\end{figure}
\renewcommand{\figurename}{Fig.}
A Test Report is defined as \texttt{<TestReport>} and is associated with the observable properties such as \texttt{<InternalIsolation>}, \texttt{<ExternalIsolation>} and \texttt{<POLVoltage>}. The metadata of the report is modeled using additional properties such as \texttt{testReportDate}, \texttt{testReportLocation}, \texttt{testReportName}, \texttt{testReportReference} and \texttt{testReportValidation}. The latter is used to indicate whether the whole Test Report is valid.

Using the ontology definition as a basis, we can streamline the extraction process. The procedure begins with parsing the Test Reports to identify relevant sections. These reports are then extracted along with their observable properties, such as POL Voltage, using the KG test table structure definition to automatically identify the purpose of each column (i.e., for the \texttt{POLVoltage} table, \texttt{Voltage Measurements [V]} contains the test data entry, while \texttt{Acceptance Limits} contains the entry validation range). Subsequently, this data is transformed into RDF triples and stored in the KG. The creation of these RDF triples is guided by the ontology, ensuring that the resulting data is both structured and machine-readable. The actual observations, which correspond to the rows in the tables, are extracted and temporarily stored in a data repository for subsequent validation.

\vspace{0.25cm}\noindent\textbf{LLM-Based Compliance Checking. }
The primary challenge in managing Test Data lies in the expensive and time-consuming task of compliance checking. This process is difficult to automate algorithmically due to the high heterogeneity in observed values and the wide variety of formats used for the acceptance limits. However, compliance checking can be automated using LLMs, as these models are capable of handling data with syntactic and structural heterogeneity. This ability makes the compliance checking process applicable across a broad spectrum of testing scenarios. Consequently, data engineers can focus only on a small subset of tests that the LLM identifies as anomalous.
The validation process is conducted row by row, rather than for the whole table at once, to prevent disclosing confidential information.
For each test result, we prompt the LLM to determine whether the measured value is within the acceptance range. The LLM’s response is then compared with the “successful” value. If there is a mismatch between these two values, the test is classified as anomalous. A test is considered valid if the measured value is within the predefined acceptance limits and the “successful” column reads “OK”, or if the value is outside the range and the “successful” column does not read “OK”.

The prompt strategy chosen is the \emph{Zero-shot}~\cite{zeroshot} (i.e., direct prompting without any examples) using a task description instead of a role-oriented approach. For data validation tasks, this strategy was shown to be superior, especially for bigger models~\cite{inconsistencies}. This is consistent with previous findings showing that zero-shot prompts are best when the task involves utilizing pre-existing knowledge embedded within the model, as opposed to learning from examples~\cite{beyondfewshots}.
Furthermore, we designed the prompt in a way that it can be applied across all types of tests and is robust to heterogeneity in the acceptance limits.
It is structured to ask a simple “True" or “False" zero-shot question that is framed as follows: \emph{“Evaluate the following electrical measure observation statement. Answer with just one “True” or “False” statement at the beginning of the answer. Is [measured\_value]  [acceptance\_limits] ?"}.
The LLM response is parsed, and the first “True” or “False” encountered is taken as the response, as sometimes the LLM might continue discussing and explaining the reasoning behind its decision.

\vspace{0.25cm}\noindent\textbf{Ontology-Based Data Access. }
We utilize a VKG to facilitate data access and manage structural heterogeneity. This VKG maps the validated Test Data storage to the ontology (refer to Figure~\ref{fig:ontology}). The knowledge within the virtualized semantic layer can be accessed via SPARQL queries, which are automatically translated into SQL.
Listing~\ref{list:simple} shows an RDF fragment modeling a POL Voltage Observation, which represents a row in the test table (refer to Figure~\ref{fig:tables}). Each row is a \texttt{sosa:Observation} with a \texttt{sosa:hasSimpleResult} value. For instance, \texttt{<http://tasi.com/pol\#TASI-1234-Core1>} is a \texttt{sosa:Observation} with a \texttt{sosa:hasSimpleResult} of “1.097 V”. This observation is associated with the \texttt{sosa:observedProperty <POLVoltage>}. The SSN ontology has been extended with two properties to accommodate the specific needs of our case study. The \texttt{tasi:reportedIn} property links the observation to the corresponding Test Report, while the \texttt{tasi:hasAcceptanceLimits} property specifies the acceptable range for the observed property. For example, \texttt{tasi:hasAcceptanceLimits "[1.076, 1.224] V"} indicates that the acceptable voltage range for the POL Voltage Observation is between 1.076V and 1.224V. The \texttt{tasi:hasTestResult} property reports the “successful” value. For instance, a successful test is indicated by \texttt{tasi:hasTestResult "OK"}.

%


\renewcommand{\thefigure}{3}
\begin{figure}[h]
\renewcommand{\figurename}{Listing}
\begin{verbatim}
mappingId  POL_Voltage_Verification
target     tasi-pol:{tr_reference}-{v_cores} a sosa:Observation ;
           rdfs:label "{tr_reference}-{v_cores}";
           sosa:observedProperty tasi:POLVoltage; 
           sosa:hasSimpleResult "{voltage_mesurements} V"^^cdt:ucum; 
           tasi:hasAcceptanceLimits {acceptance_limits}; 
           tasi:testReportDate {test_report_date}^^xsd:dateTime;
           tasi:hasTestResult {successful};
           tasi:reportedIn {tr_reference}. 
source     SELECT tr_reference, v_cores, voltage_mesurements
           acceptance_limits, test_report_date, successful
           FROM tasi.pol_voltage    
\end{verbatim}
\caption{The mapping for the \texttt{POLVoltage Observation}.}
\label{list:polmap}
\end{figure}
\renewcommand{\figurename}{Fig.}

To populate the ontology, we establish a series of mappings. These mappings create connections between the ontology and the underlying data storage, thereby providing semantic meaning to the Test Data. An example of mapping for a \texttt{POLVoltageObservation} is provided in Listing~\ref{list:polmap}.
The mapping is defined with a \texttt{mappingId} of \texttt{POL Voltage Verification}, which corresponds to the type of test being performed. The \texttt{target} of the mapping is a URI that represents a \texttt{sosa:Observation} in the ontology.
The \texttt{source} is a SQL query that retrieves the necessary data from the \texttt{POL Voltage Verification} table in the test results storage. The variables in the \texttt{source} query correspond to the placeholders in the \texttt{target}. Once the mapping is executed, these placeholders are replaced with the actual values retrieved by the \texttt{source} query. This allows us to virtually represent the storage as an RDF graph, integrating different data sources into a unified view.

\section{Implementation and Evaluation}
\label{impl}
In this section, we delve into the specifics of our system’s implementation and the technologies used. Following this, we present a benchmarking study of various state-of-the-art LLMs to evaluate their capability of performing automated compliance checking. An evaluation of the whole methodology from a cost-benefit perspective is provided in Section~\ref{lessons}.

\vspace{0.25cm}\noindent\textbf{Implementation details. }
An \emph{Apache Airflow} DAG (Directed Acyclic Graph) was designed to orchestrate the entire process. Apache Airflow is a popular open-source tool for creating, scheduling, and monitoring data pipelines. For modeling the Test Reports and their properties, we implemented the KG using \emph{Apache Jena Fuseki}, a server that allows for querying and updating the KG using the SPARQL query language. The test results are stored in an \emph{Apache Parquet}, a free and open-source column-oriented data storage, which allows handling large volumes of data while maintaining high performance.
The VKG was implemented using \emph{OntopSpark}~\cite{ontopspark}, an extension developed by Politecnico di Milano of \emph{Ontop}~\cite{ontop}, an open-source OBDA system that allows for querying relational data sources through an ontology via R2RML~\cite{r2rml} mappings. We do not report a detailed analysis of Ontop performances since it was benchmarked in several other papers~\cite{ontopbench1,ontopbench2}. We present a discussion about the effort to solve the problem with and without our solution in Section~\ref{lessons}.

\begin{table}[t]
    \caption{The results of the comparative analysis on state-of-the-art LLMs for compliance checking, highlighting the superior performance of GPT-4 and Gemini Ultra.} 
    \centering 
    \label{tab:benchmarking} 
    \begin{tabularx}{\textwidth}{lXccccc} 
        \toprule 
        \textbf{Model} & \textbf{Test Type} & \textbf{\#Tests} & \textbf{Accuracy} & \textbf{Precision} & \textbf{Recall} & \textbf{F1-Score} \\ 
        \midrule 
        \multirow{4}{*}{{\textit{GPT-3.5}}} 
            & POL Voltage & 53 & 0.868 & 0.625 & 0.556 & 0.588 \\ 
            & Internal Isolation & 86 & 0.779 & 0.056 & 0.333 & 0.095 \\ 
            & External Isolation & 59 & 0.932 & 0.333 & 0.333 & 0.333 \\
            & \textbf{Overall} & 198 & 0.849 & 0.241 & 0.467 & 0.318 \\ 
        \midrule 
        \multirow{4}{*}{{\textit{GPT-4}}} 
            & POL Voltage & 53 & 0.981 & 1.000 & 0.900 & 0.947 \\ 
            & Internal Isolation & 86 & 1.000 & 1.000 & 1.000 & 1.000 \\  
            & External Isolation & 59 & 0.983 & 0.750 & 1.000 & 0.857 \\
            & \textbf{Overall} & 198 & \textbf{0.990} & \textbf{0.938} & 0.938 & \textbf{0.938} \\ 
        \midrule 
        \multirow{4}{*}{{\textit{Gemini Ultra}}} 
            & POL Voltage & 53 & 1.000 & 1.000 & 1.000 & 1.000 \\ 
            & Internal Isolation & 86 & 1.000 & 1.000 & 1.000 & 1.000 \\ 
            & External Isolation & 59 & 0.949 & 0.500 & 1.000 & 0.667 \\
            & \textbf{Overall} & 198 & 0.985 & 0.833 & \textbf{1.000} & 0.909 \\
        \midrule 
        \multirow{4}{*}{{\textit{Mixtral 8x7B}}} 
            & POL Voltage & 53 & 0.925 & 0.875 & 0.700 & 0.778 \\ 
            & Internal Isolation & 86 & 0.663 & 0.150 & 0.200 & 0.171 \\ 
            & External Isolation & 59 & 0.644 & 0.136 & 0.600 & 0.222 \\
            & \textbf{Overall} & 198 & 0.727 & 0.260 & 0.433 & 0.325 \\ 
        \midrule 
        \multirow{4}{*}{{\textit{LLama 2 70B }}} 
            & POL Voltage & 53 & 0.887 & 0.800 & 0.444 & 0.571 \\ 
            & Internal Isolation & 86 & 0.733 & 0.091 & 0.400 & 0.148 \\ 
            & External Isolation & 59 & 0.712 & 0.111 & 0.667 & 0.191 \\
            & \textbf{Overall} & 198 & 0.768 & 0.178 & 0.471 & 0.258 \\ 
        \midrule 
        \multirow{4}{*}{{\textit{Claude 3 Opus }}} 
            & POL Voltage & 53 & 1.000 & 1.000 & 1.000 & 1.000 \\ 
            & Internal Isolation & 86 & 0.895 & 0.250 & 1.000 & 0.400 \\ 
            & External Isolation & 59 & 0.983 & 0.750 & 1.000 & 0.857 \\
            & \textbf{Overall} & 198 & 0.949 & 0.615 & \textbf{1.000} & 0.762 \\ 
        \bottomrule 
    \end{tabularx} 
\end{table}

\vspace{0.25cm}\noindent\textbf{LLMs Benchmarking. }
A benchmarking study was carried out to assess the performance of state-of-the-art LLMs in automated compliance checking. The models tested included GPT-3.5~\cite{gpt3}, GPT-4~\cite{gpt4}, Gemini Ultra~\cite{gemini}, Mixtral 8x7B~\cite{mixtral}, LLama 2 70B~\cite{llama2}, and Claude 3 Opus~\cite{claude3}. Performance was measured using standard metrics such as accuracy, precision, recall, and F1-score. The positive class was considered when the measured values fell outside the acceptance limit range, which is also the less represented class. The models were tested across three test categories: POL Voltage Verification, Internal Isolation, and External Isolation.

Table~\ref{tab:benchmarking} presents the benchmarking results, showing a clear distinction in performance among the tested LLMs. GPT-4 and Gemini Ultra are the top performers across all test categories, with GPT-4 achieving the highest overall accuracy, precision, and F1-score. Gemini Ultra, on the other hand, achieved perfect scores in the POL Voltage and Internal Isolation tests and had the highest overall recall. In contrast, GPT-3.5, Mixtral 8x7B, and LLama 2 70B consistently underperformed compared to the top models, rendering them unsuitable for our task. Claude 3 Opus demonstrated strong performance in the POL Voltage test but had lower precision in the Internal Isolation and External Isolation tests due to its inability to handle cases where the test result’s unit of measure was absent, a frequent scenario in the Isolation tests. This benchmarking study provides strong evidence supporting the effectiveness of an LLM-based approach for automated compliance checking using state-of-the-art LLMs. The performance of GPT-4 and Gemini Ultra underscores the potential of these models in managing complex data validation tasks.

\section{Discussion on Uptake and Lessons Learned}
\label{lessons}

\noindent\textbf{Benefits and Scalability. }
The transition from the current method to the proposed solution suggests a significant reduction in the effort measured in person-days required to complete and validate Test Reports before extracting longitudinal Test Data to analyze. The existing procedure necessitates substantial manual work for tasks such as creating Test Report templates, instantiating Test Reports, filling in the test results and checking the compliance with the acceptance limits, reviewing the Test Data and their coherence with the reported success, looking for errors and correcting them, and extract/transform data to perform longitudinal analysis (see Figure~\ref{fig:pipeline_as_is}). The proposed solution, while requiring the modeling and maintenance of an ontology that encapsulates various test types (refer to Figure~\ref{fig:heatmap}), and the annotation of the template with semantic tags that define each section, is fully automated (see Figure~\ref{fig:pipeline}). This includes the extraction of test results and acceptance limits, error isolation, requests for manual review and correction, and data accessibility for longitudinal analysis. 

\renewcommand{\thefigure}{6}
\begin{figure}
    \centering
    \includegraphics[width=1\textwidth]{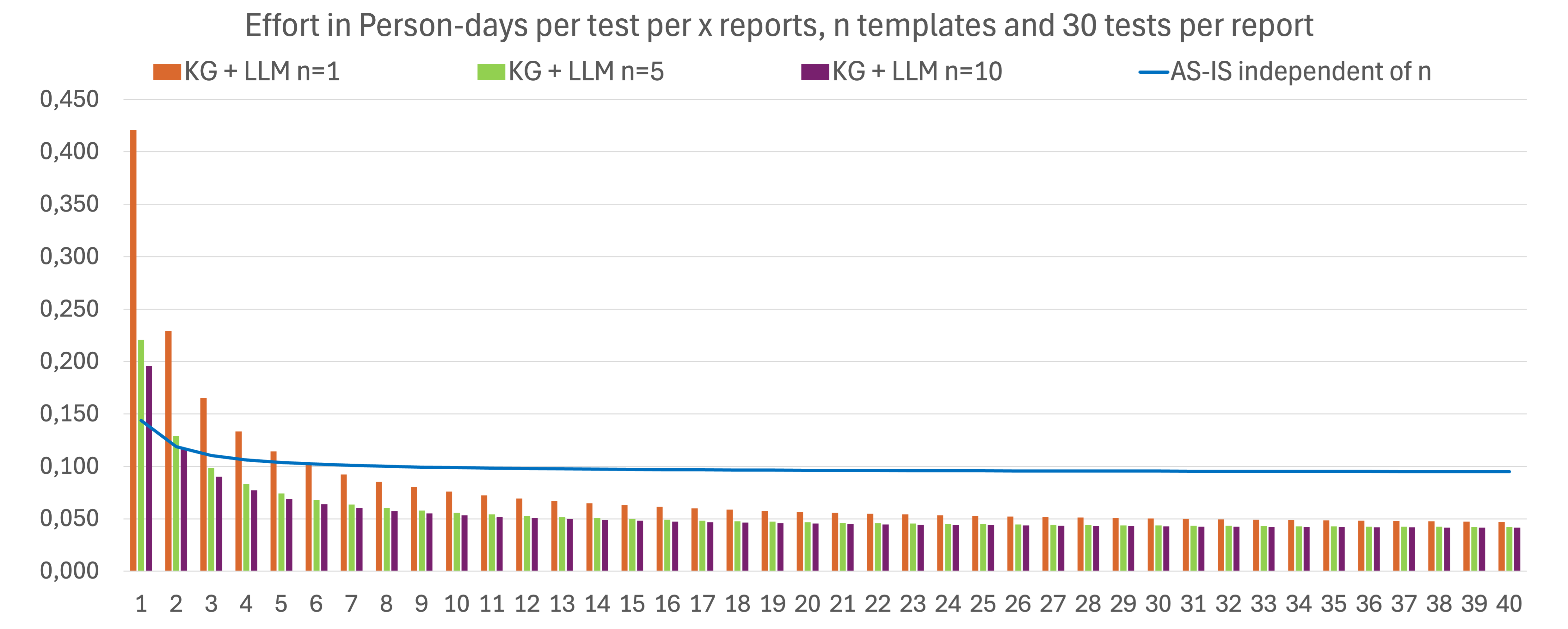}
    \hfill
    \includegraphics[width=1\textwidth]{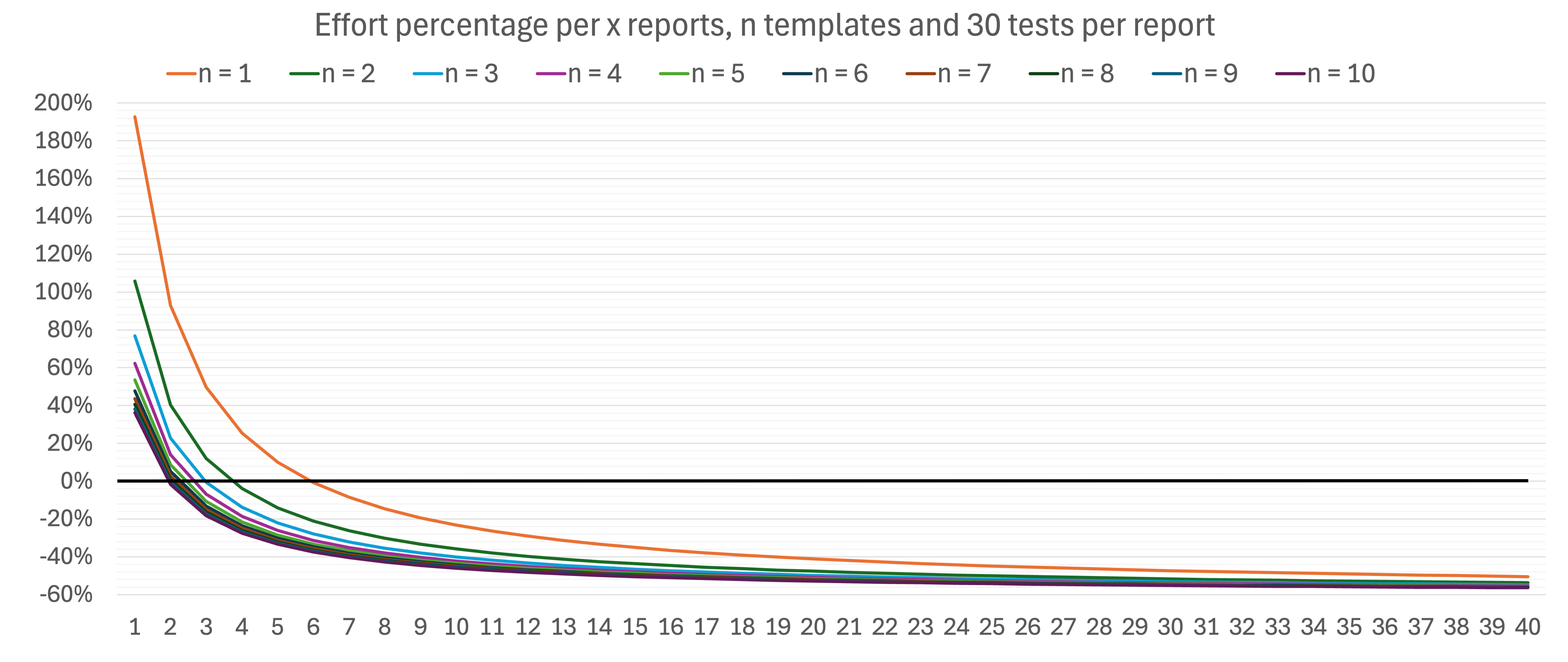}
    \caption{Comparison of effort reduction between our solution (KG+LLM) and the current method (AS-IS). The effort depends on the number of templates ($n$), the number of reports (x-axis), and the test types per report which for simplicity are set to 30.}
    \label{fig:cost-model-1-template}
\end{figure}

We developed a cost model based on the experience documented in this paper.
This model estimates the effort involved as a product of three factors: the number of different Test Report templates, the average number of Test Reports per template, and the number of test types (e.g., POL Voltage Verification, Preliminary Power Consumption) per report.
Comparing the effort required to model, compile, and validate from 1 to 10 Test Report templates, each with an average of 30 types of test per report (refer to Figure~\ref{fig:cost-model-1-template}), we derive that as the number of reports (on the x-axis) increases:
\begin{itemize}
    \item For a single Test Report template (n=1), benefits start to appear after the 6th report.
    \item For five templates (n=5), benefits are seen after the 3rd report.
    \item For ten templates (n=10), the benefits are obtained at the 2nd report.
\end{itemize}

As the number of Test Report templates ($n$) increases, the number of Test Reports (on the x-axis) needed to see the benefits of using KG and LLMs is significantly reduced, with potential time savings of more than 50\%. The right part of Figure~\ref{fig:cost-model-1-template} illustrates the break-even points for an increasing number of Test Reports in detail.

\vspace{0.25cm}\noindent\textbf{Next Steps for Large Scale Deployment. }
The proof-of-concept of the proposed solution was well received by Thales Alenia Space, but additional efforts are needed for the transition to a large-scale deployment.
We are currently engaged in a feasibility study to port the solution to the other five nations in which Thales Alenia Space operates (France, Belgium, Spain, Switzerland, and the UK). 
Since the scenarios can vary significantly in these different divisions, this can potentially broaden adoption across the aerospace industry through further development and demonstration of value in diverse operational environments.
Furthermore, despite the proposed solution focusing on a specialized case study, the principle of data validation via LLMs, to simplify the conceptual modeling process and reduce manual work, could potentially be extended to other scenarios such as mechanical and electrical qualification, given the ability of KGs and LLMs to adapt to different tasks and data types. However, it’s true that to apply this approach in different contexts, slight reconfigurations would be necessary. Furthermore, it would be essential to establish benchmarks for each specific use case to evaluate the applicability in a new scenario.
Given the significant savings, Thales Alenia Space expresses its intention to continue prototyping for other types of tests on PCBs, extend to the other product lines, and eventually deploy to all product lines. Preliminary experiments in this direction have already produced some promising results.

\vspace{0.25cm}\noindent\textbf{Lessons Learned. }
The development of the proposed solution revealed several key lessons. Firstly, the success of the implementation heavily relied on a well-structured ontology and clean mappings. The initial investment in modeling proved beneficial, as it minimized downstream efforts. Additionally, the integration of LLMs streamlined data validation, drastically reducing the need for manual intervention.
Identified best practices include the necessity for iterative development and validation of the ontology and its corresponding mappings. This ensures accurate modeling of test template reports. Moreover, it is crucial to conduct a comparative evaluation of alternative LLMs to stay updated with the evolving heterogeneities in Test Data, acceptance limits, and report requirements. Collaboration with stakeholders and domain experts was also essential for fine-tuning the KG and LLM prompts for optimal performance, and ensuring that confidentiality requirements were met while incorporating closed-source LLMs in the pipeline.
As we move forward, these insights will guide our efforts to extend the solution to other product lines and further enhance the system's performance and reliability.

\section{Conclusion and Future Work}
\label{conclusions}
In this paper, we demonstrated a successful application of Semantic Web technologies combined with LLMs for integrating and validating heterogeneous and unstructured industrial data through a use case related to Test Reports of electronic boards used in Thales Alenia Space's satellite systems. Our benchmarking study revealed that GPT-4 and Google Gemini possess remarkable abilities in automating the process of compliance checking. Considering that LLMs are still in the early stages of their development, it’s reasonable to expect their performance to improve further, enabling them to handle even more complex data validation tasks in the near future. 
Overall, the proposed solution demonstrates a clear cost-benefit advantage over the existing document-centric solution. The potential efficiency gains underscore the value of investing in advanced AI-driven automation for such data-intensive tasks.

As future work, we intend to investigate whether the use of LLMs can be extended to perform automatic ontology construction, utilizing document tags, and also data homogenization. This would involve parsing the test results and the acceptance limits through an LLM-based approach. Additionally, we aim to further enhance data access by employing LLMs to convert natural language into SPARQL queries, thereby enabling Thales Alenia Space engineers to access knowledge directly using natural language.

\paragraph*{Supplemental Material Statement:} The obfuscated dataset and the code used to benchmark the Large Language Models are made available for enhancing the reproducibility of the study and potential reuse for future research. However, please note that the code and data related to other steps of the methodology are not provided as they comprise confidential information belonging to Thales Alenia Space. The available resources can be accessed in our GitHub Repository at \url{https://github.com/Antonio-Dee/tasi-testdata}.

\small\vspace{0.25cm}\noindent\textbf{Acknowledgments.} Antonio De Santis's doctoral scholarship is funded by the Italian Ministry of University and Research (MUR) under the National Recovery and Resilience Plan (NRRP), by Thales Alenia Space, and by the European Union (EU) under the NextGenerationEU project.
%
%
%
\bibliographystyle{splncs04}
\bibliography{tasi}

\end{document}